%% file: root_arxiv.tex
\documentclass{article}

\usepackage{xcolor}    
\usepackage[preprint]{corl_2026} 

\usepackage{amsmath,amssymb,amsfonts}
\usepackage{bm}

\usepackage{graphicx}
\usepackage{booktabs}
\usepackage{multirow}
\usepackage{subcaption}
\usepackage{xcolor}

\usepackage{tikz}
\usepackage{pgfplots}
\usetikzlibrary{arrows.meta,positioning,shapes.geometric,calc,backgrounds,fit}
\pgfplotsset{compat=1.18}

\usepackage{algorithm}
\usepackage{algpseudocode}

\usepackage{enumitem}
\setlist[itemize]{leftmargin=1.5em, itemsep=2pt, topsep=2pt, parsep=0pt}
\setlist[enumerate]{leftmargin=1.8em, itemsep=2pt, topsep=2pt, parsep=0pt}

\definecolor{simblue}{RGB}{70,130,180}
\definecolor{methodgreen}{RGB}{60,160,80}
\definecolor{realred}{RGB}{190,70,70}
\definecolor{policypurple}{RGB}{140,90,180}
\definecolor{curricorange}{RGB}{220,140,40}

\title{Pose-Agnostic Robotic Functional Grasping via Observation-Action Canonicalization}

\author{
  Le Qiu$^{1}$ \quad Cole Harrison$^{2}$ \quad Jiankai Sun$^{3}$ \quad Yao Liu$^{4}$ \\
  \textbf{Suning Huang$^{3}$ \quad Qianzhong Chen$^{3}$ \quad Yang You$^{3, \dagger}$ \quad Marco Pavone$^{3}$}\\[2pt]
  $^{1}$Tsinghua University\quad
  $^{2}$Amazon\quad
  $^{3}$Stanford University\quad
  $^{4}$University of California, Berkeley
}

\begin{document}
\maketitle

\makeatletter
\renewcommand{\@makefntext}[1]{\noindent #1}
\makeatother
\footnotetext[0]{\textsuperscript{$\dagger$} For any questions, please contact: \texttt{yangyou@stanford.edu}}

\begin{abstract}
\input{sections/00abs}
\end{abstract}


\input{sections/10intro}

\input{sections/20related}

\input{sections/30pre}

\input{sections/40method}

\input{sections/50exp}

\input{sections/60limit}

\bibliography{Reference}
\newpage
\appendix
\section*{Appendix}
\input{sections/70appen}

\end{document}

%% file: sections/00abs.tex
Functional robotic grasping requires a policy that generalizes across diverse object geometries and poses while maintaining task-specific contact precision.
We study this challenge through mug-handle grasping, where thin handles, instance variation, and upright or inverted placements make both perception and control sensitive to object configuration.
Grasp pose detection methods operate open-loop and are sensitive to estimation errors on thin handle structures. Learned visuomotor policies must implicitly learn to handle the coupled variation in visual appearance and action direction induced by different object placements, limiting generalization. We propose \textbf{AnyMug}, a canonicalized visuomotor reinforcement learning framework for functional grasping that trains a single closed-loop policy entirely in simulation and deploys it zero-shot on a real robot.
AnyMug introduces observation-action canonicalization, which transforms both the depth observation and the predicted end-effector action into a shared object-centric frame. The policy therefore sees a consistent mug-centered view and emits actions in a canonical direction regardless of mug placement, allowing the same grasping behavior to be reused across configurations.
A handle-aware reward further encourages precise approach, gripper alignment, and opposing-finger placement, while a pose curriculum and domain randomization improve training stability and sim-to-real transfer.
In simulation, AnyMug achieves over 93\% success rate on both unseen upright and inverted mugs and transfers zero-shot to a real Franka Panda, reaching 80\% success rate on 5 held-out physical mugs across both pose categories.

%% file: sections/10intro.tex
\section{Introduction}
\label{sec:intro}

Robotic grasping is a fundamental problem in manipulation~\citep{bicchi2000robotic, bohg2013data,sun2024arch,chen2020transferable,chen2026sarm2,huang2026breaking,huang2025spatial,wang2023mimicplay,huang2025particleformer}, yet functional grasping remains difficult when objects afford only specific useful contacts. Mugs are a representative example where successful manipulation often requires grasping the handle rather than the body. Handles are thin, curved, and vary substantially across instances. Moreover, the same mug may appear at different positions, handle orientations, and in two pose categories: upright and inverted. A robust policy must therefore generalize across both geometric variation across instances and object pose while maintaining precise gripper alignment and closure timing.

Existing grasping methods can be broadly categorized into two strategies. Grasp pose detection methods~\citep{mahler2017dex, mousavian20196, fang2023anygrasp} predict a target 6-DoF or 7-DoF grasp pose from visual observations, but execution is typically open-loop and sensitive to small pose estimation errors on thin handle structures~\citep{qi2017pointnet, you2024cppf++}.
Closed-loop visuomotor policies~\citep{levine2018learning, dmitry2018qt, chi2025diffusion, zitkovich2023rt} can adapt during execution, but typically learn placement and orientation generalization implicitly from data. For mug-handle grasping, this requires learning many placement-dependent approach directions and closure timings, even when the underlying grasp strategy is geometrically equivalent across placements.

\begin{figure*}[t]
\centering
\includegraphics[width=\textwidth]{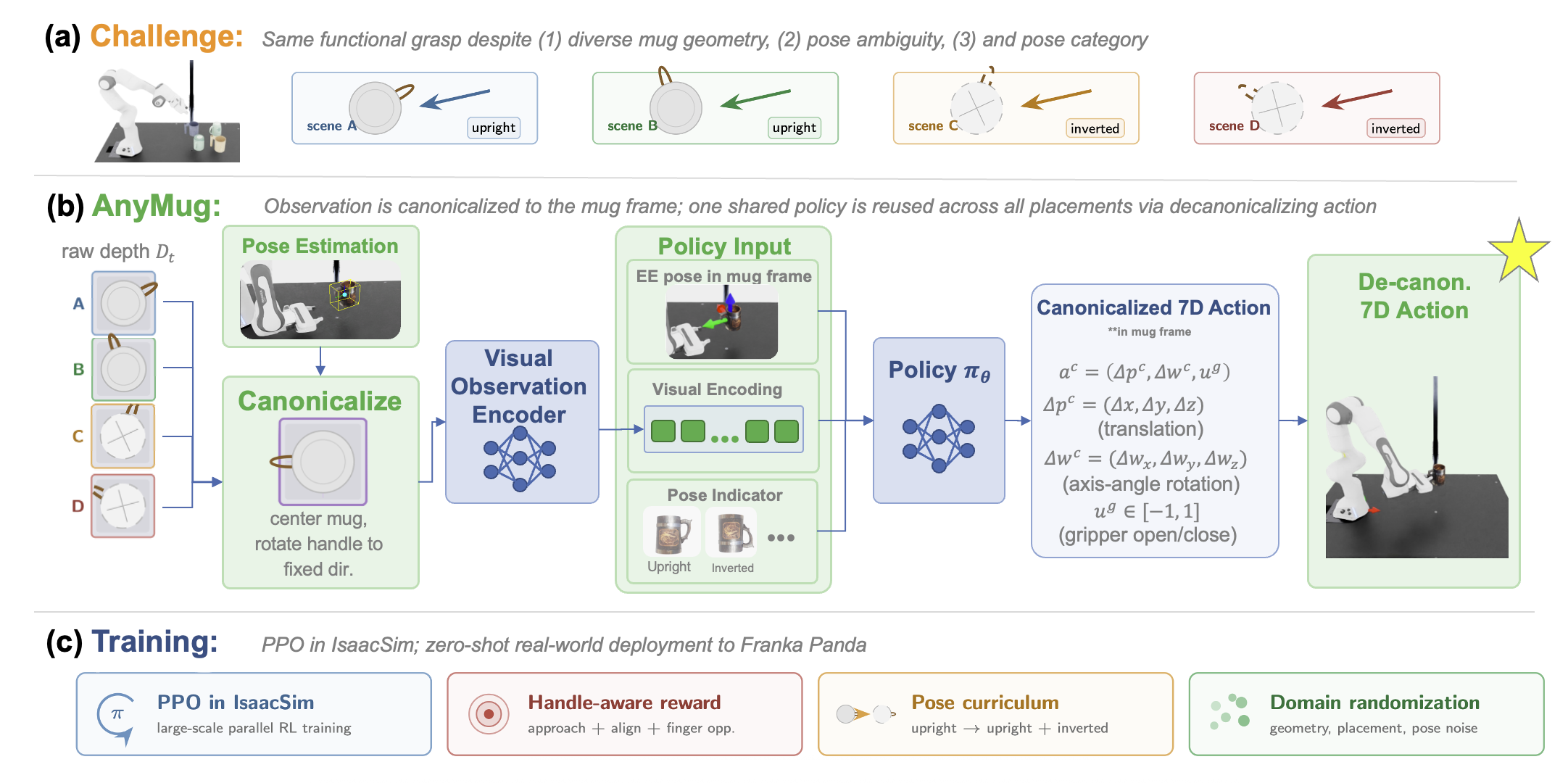}
\caption{\textbf{Overview of AnyMug.} \textbf{(a) Challenge:} Functional mug-handle grasping must generalize across mug geometries, placements, and pose categories. \textbf{(b) AnyMug:} AnyMug canonicalizes the top-view depth image $D_t$ into a mug-centric frame by centering the mug and aligning the handle to a fixed direction. A shared policy $\pi_\theta$ takes the canonical depth, mug-frame end-effector pose, and pose indicator, and predicts a canonical 7D action $a^c=(\Delta p^c,\Delta \omega^c,u^g)$, whose spatial components are de-canonicalized to the world frame for execution. \textbf{(c) Training:} The policy is trained with PPO using handle-aware rewards, pose curriculum learning, and domain randomization, enabling zero-shot deployment on a real Franka Panda.}
\label{fig:overview}
\vspace{-2.0em}
\end{figure*}

We present \textbf{AnyMug}, a reinforcement learning framework for pose-agnostic functional mug-handle grasping. We propose observation-action canonicalization. We warp the top-view depth observation so that the mug is centered and the handle appears in a canonical direction, and represent the end-effector action in the same mug-centric frame. The predicted action is then de-canonicalized back to the world frame for execution. This removes pose-induced variation from both perception and control, enabling a single policy to reuse a shared closed-loop grasping strategy across mug positions and handle orientations.

Canonicalization provides a consistent object-centric representation, but the grasp execution still depends on instance-specific handle geometry, local depth cues, and residual pose-estimation error. We therefore train a handle-aware closed-loop reinforcement learning (RL) policy that learns instance-level adaptation from canonicalized depth and proprioceptive feedback. Dense geometric rewards guide handle approach, gripper alignment, and opposing-finger placement before closure. This encourages precise gripper-handle alignment and gripper-closing timing.

We evaluate AnyMug in simulation and zero-shot on a real Franka Panda robot. The policy is trained entirely in Isaac Sim~\citep{makoviychuk2021isaac} and tested on unseen mug instances and diverse pose configurations. Empirically, AnyMug achieves over 93\% success rate in simulation and 80\% success rate on 5 unseen physical mugs across both pose categories, while ablations confirm that observation-action canonicalization and the finger-opposition reward are the dominant contributors. In summary, our contributions are:
\begin{itemize}
    \item \textbf{Observation-action canonicalization.}
    We formulate mug grasping as a canonicalized problem where both depth observations and end-effector actions are represented in a shared mug-centric frame, reducing pose-induced variation in perception and control.
    
    \item \textbf{Closed-loop handle-aware grasping.}
    We train a single RL policy with canonicalized visual-proprioceptive feedback and handle-aware rewards, enabling precise approach, alignment, and closure across diverse mug geometries and pose configurations.

    \item \textbf{Comprehensive evaluation and analysis.}
    We benchmark AnyMug against motion planning and Diffusion Policy in simulation and on hardware, using calibrated pose-estimation noise and a $5$-mug zero-shot real-robot test set. Ablations further isolate the contribution of canonicalization, reward shaping, and the pose curriculum.

\end{itemize}

%% file: sections/20related.tex
\section{Related Work}
\label{sec:related}

We discuss prior work on grasp pose detection, closed-loop vision-based grasping, and canonical or equivariant representations for manipulation in Appendix~\ref{apx:related}. The closest paradigms either canonicalize only observations or constrain the policy network while leaving actions in the robot frame, requiring the policy to learn placement-dependent motor commands. AnyMug differs by jointly canonicalizing both the top-view depth observation and the end-effector action into a shared mug-centric frame, enabling a single closed-loop policy to reuse the same grasping strategy across mug poses and handle orientations.

%% file: sections/30pre.tex
\section{Preliminaries}
\label{sec:pre}

\subsection{Problem Formulation}

We formulate pose-agnostic mug grasping as a finite-horizon partially observable Markov Decision Process (POMDP) $\mathcal{M} = (\mathcal{S}, \mathcal{A}, \mathcal{O}, \mathcal{P}, \mathcal{R}, \gamma, H)$, where $\mathcal{S}$ is the state space, $\mathcal{A}$ is the action space, $\mathcal{O}$ is the observation space, $\mathcal{P}$ is the transition probability function, $\mathcal{R}$ is the reward function, $\gamma \in (0,1)$ is the discount factor, and $H$ is the episode horizon. At each timestep $t$, the policy receives an observation $o_t \in \mathcal{O}$ and takes an action according to $\pi_\theta(a_t \mid o_t)$, where $\theta$ denotes the policy parameters. The environment transitions to state $s_{t+1} \sim \mathcal{P}(\cdot \mid s_t, a_t)$ and returns reward $r_t = \mathcal{R}(s_t, a_t)$. The objective is to maximize the discounted cumulative rewards $J(\pi_\theta) = \mathbb{E}_{\tau \sim \pi_\theta} \left[ \sum_{t=0}^{H-1} \gamma^t r_t \right]$, where $\tau = \{(s_t, a_t, r_t)\}_{t}$ denotes a sequence of state-action-reward triplets.

\textbf{Observation Space.}
The policy observation is $o_t = (D_t,\, p_{\mathrm{rel}}, \, q_{\mathrm{rel}}, \, c)$, where $D_t \in \mathbb{R}^{H_d \times W_d}$ is a top-view depth image, $p_{\mathrm{rel}} \in \mathbb{R}^3$ and $q_{\mathrm{rel}} \in \mathbb{R}^4$ are the end-effector position and orientation expressed in the mug-centric frame, and $c \in \{0,1\}^{K}$ is a one-hot pose-category indicator over $K$ categories. In our experiments, $H_d = W_d = 120$ and $K=2$ for upright and inverted mugs. The canonicalization of $D_t$ is described in Sec.~\ref{sec:canon}.

\textbf{Action Space.}
The policy outputs a continuous 7D end-effector command $a_t^c = (\Delta p_t^c, \Delta \omega_t^c, u_t^g)$ in the canonical frame, where $\Delta p_t^c \in \mathbb{R}^3$ is a translational displacement, $\Delta \omega_t^c \in \mathbb{R}^3$ is an axis-angle rotational displacement, and $u_t^g \in \mathbb{R}$ is the gripper command. The gripper closes when $u_t^g>0.05$, opens when $u_t^g<-0.05$, and remains unchanged when $|u_t^g|\leq0.05$ to avoid spurious actuation. The canonical action is then de-canonicalized to the world frame as described in Sec.~\ref{sec:canon}.

\subsection{Task Objective}
Our objective is to learn a single closed-loop policy $\pi_\theta: \mathcal{O} \to \mathcal{A}$ that achieves functional handle grasping across diverse mug instances and pose categories. A successful episode requires the robot to approach the handle, align the gripper, close around the handle, and lift the mug without tipping or dropping it. We optimize $\pi_\theta$ with PPO~\citep{schulman2017proximal} in simulation.

%% file: sections/40method.tex
\section{Methodology}
\label{sec:method}

We present AnyMug, a reinforcement learning framework for functional mug-handle grasping that trains a single closed-loop policy generalizing across diverse mug instances and pose categories, and transfers zero-shot to a real robot. AnyMug first canonicalizes both the top-view depth observation and the end-effector action into a mug-centric frame to reduce placement variation while preserving task-relevant geometry. Then AnyMug trains the policy with a handle-aware reward that provides dense guidance for handle approach, gripper alignment, and closure timing. To further improve training stability and sim-to-real robustness, AnyMug combines pose-category curriculum learning and domain randomization. The overview of our method is shown in Fig.~\ref{fig:overview}.

\subsection{Observation-Action Canonicalization}
\label{sec:canon}

\begin{figure}[t]
\centering
\includegraphics[width=\columnwidth]{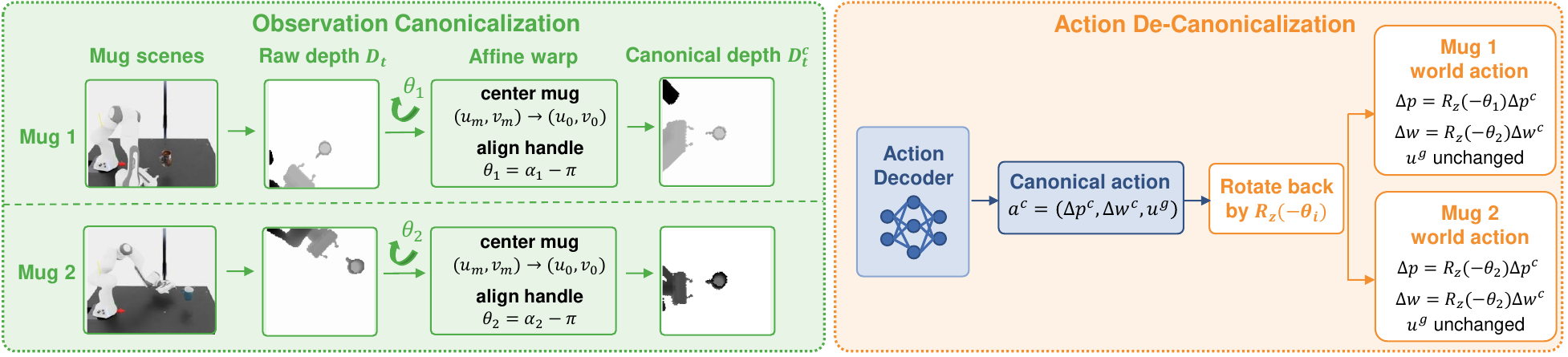}
\caption{\textbf{Observation-action canonicalization.} Left: We canonicalize each top-view depth observation $D_t$ by projecting the mug center and handle center into the image, estimating the handle direction $\alpha_t$, and applying an affine warp that centers the mug and aligns the handle to a fixed canonical direction. This produces a canonical depth observation $D_t^c$. Right: The policy predicts a canonical action $a_t^c=(\Delta p_t^c, \Delta \omega_t^c, u_t^g)$ in the same mug-centric frame. Before execution, the spatial components are rotated back to the world frame using $R_z(-\theta_t)$. This canonical frame reduces pose-induced variation in both perception and control while preserving mug-specific handle geometry.}
\label{fig:canon}
\end{figure}

Raw top-view depth images vary substantially with mug position and handle yaw, although the desired handle-grasping behavior is geometrically equivalent in a mug-centric frame. To reduce this variance, we canonicalize both the depth observation and action representation using a shared mug-centric alignment, as shown in Fig.~\ref{fig:canon}. Superscripts $w$ and $c$ denote world frame and canonical frame quantities respectively.

\textbf{Observation Canonicalization.}
Let $D_t \in \mathbb{R}^{H_d \times W_d}$ denote the top-view depth image captured at timestep $t$, and let $p_m^w \in \mathbb{R}^3$ and $p_h^w \in \mathbb{R}^3$ denote the 3D positions of the mug center and handle center in the world frame. Using the known camera intrinsic matrix $K \in \mathbb{R}^{3\times3}$ and the camera extrinsic transform, we first project both points into the image coordinates:
\begin{equation}
    (u_m, v_m) = \Pi(p_m^w), \quad (u_h, v_h) = \Pi(p_h^w),
\end{equation}
where $\Pi: \mathbb{R}^3 \to \mathbb{R}^2$ denotes the projection. The handle direction in the image coordinates is $\alpha_t = \operatorname{atan2}(v_h - v_m,\; u_h - u_m)$. We define the canonical handle direction as $\alpha^\star=\pi$, corresponding to a left-facing handle in the canonical image. The yaw rotation required to align the observed handle with this convention is: $\theta_t =
    \operatorname{atan2}\!\left(\sin(\alpha_t-\alpha^\star), \cos(\alpha_t-\alpha^\star)\right)$.
The canonical depth image $D_t^c$ is obtained by an affine warp $\mathcal{W}$ that simultaneously centers the mug and aligns the handle:
\begin{equation}
    D_t^c = \mathcal{W}(D_t;\, A_t), \quad
    A_t =
    \begin{bmatrix}
        \cos\theta_t & -\sin\theta_t & -\tau_x \\[2pt]
        \sin\theta_t &  \cos\theta_t & -\tau_y
    \end{bmatrix}
    \in \mathbb{R}^{2\times3},
    \label{eq:affine_warp}
\end{equation}
where $\tau_x = 2(u_0 - u_m)/(W_d{-}1)$ and $\tau_y = 2(v_0 - v_m)/(H_d{-}1)$ are normalized translations that move the projected mug center $(u_m, v_m)$ to the image center $(u_0, v_0) = ((W_d{-}1)/2,\,(H_d{-}1)/2)$. Thus, $D_t^c$ presents the mug at a consistent image location and handle orientation regardless of the mug's raw physical placement.

\textbf{Action De-canonicalization.} The policy predicts a canonical end-effector action $a_t^c = (\Delta p_t^c,\, \Delta\omega_t^c,\, u_t^g) \in \mathbb{R}^7$, where $\Delta p_t^c \in \mathbb{R}^3$ is a translational displacement, $\Delta\omega_t^c \in \mathbb{R}^3$ is an axis-angle rotational displacement, and $u_t^g \in \mathbb{R}$ is the gripper command for close or open. Before execution, we undo the canonical rotation to map the spatial action back to the world frame:
\begin{equation}
    \Delta p_t^w = R_z(-\theta_t)\,\Delta p_t^c, \qquad
    \Delta\omega_t^w = R_z(-\theta_t)\,\Delta\omega_t^c,
    \label{eq:action_decanon}
\end{equation}
where $R_z(\phi) \in \mathrm{SO}(3)$ denotes a rotation by angle $\phi$ about the world vertical $z$-axis. The gripper command is not rotated and is executed directly. The resulting world-frame command $(\Delta p_t^w,\Delta\omega_t^w,u_t^g)$ is sent to a differential inverse-kinematics controller, which converts the desired end-effector motion into joint-position targets.

The canonical frame is recomputed at every control step from the current estimated mug center and handle direction, ensuring that the depth warp and action de-canonicalization remain consistent with small mug motions during closed-loop interaction.

\subsection{Handle-Aware Reward}
\label{sec:reward}
Functional handle grasping requires the gripper to approach the handle with the correct orientation and close only after the fingers are positioned on opposite sides of the handle. To achieve this, we use a handle-aware reward that provides dense geometric guidance for approach, alignment, and pre-closure finger placement: 
\begin{equation}
    r_t =
    w_p\, r_{\mathrm{pos}}
    + w_R\, r_{\mathrm{rot}}
    + w_o\, r_{\mathrm{opp}}
    - w_a\, r_{\mathrm{act}}.
    \label{eq:reward_total}
\end{equation}
where $w_p,w_R,w_o,w_a>0$ are scalar weights.

\textbf{Approach and Orientation.}
Let $p_g^w \in \mathbb{R}^3$ and $q_g^w \in \mathbb{S}^3$ denote the gripper grasp point and orientation in the world frame, and let $p_h^w \in \mathbb{R}^3$ denote the handle center. We use a standard reaching reward 
$r_{\mathrm{pos}} = \exp\!\left(-\lambda_p \|p_g^w-p_h^w\|_2\right)$,
where $\lambda_p>0$ controls the distance sensitivity. 
To encourage a feasible handle-grasp orientation, we define a pose-dependent target orientation $q_h^w = q_m^w \otimes q_{\mathrm{rel}}(c)$,
where $q_m^w$ is the mug orientation, $c$ is the pose-category indicator, and $q_{\mathrm{rel}}(c)$ specifies a canonical gripper orientation relative to the mug for each category. The orientation reward is 
$r_{\mathrm{rot}} = \exp\!\left(-\lambda_R d_q(q_g^w,q_h^w)\right)$,
where $\lambda_R>0$ and $d_q(\cdot,\cdot)$ denotes quaternion geodesic distance.

\textbf{Finger Opposition Reward.}
The key task-specific term encourages the two fingertips to approach the handle from \emph{opposite sides} before closure. Let $p_l^w, p_r^w$ denote the left and right fingertip positions. We define normalized handle-to-fingertip directions:
\begin{equation}
    \hat{v}_l = \frac{p_l^w - p_h^w}{\|p_l^w - p_h^w\| + \epsilon}, \qquad
    \hat{v}_r = \frac{p_r^w - p_h^w}{\|p_r^w - p_h^w\| + \epsilon},
\end{equation}
where $\epsilon>0$ is a small constant for numerical stability. 
The opposition score $a_{\mathrm{opp}} = -\hat{v}_l^\top \hat{v}_r$ is maximized when the fingertips are on opposite sides of the handle. To avoid rewarding fingers that are opposite but far from the handle, we gate this score by fingertip proximity:
\begin{equation}
    r_{\mathrm{opp}} = a_{\mathrm{opp}} \cdot \exp\!\bigl(-\lambda_f\, d_{\max}\bigr), \qquad 
    d_{\max} = \max(\|p_l^w - p_h^w\|,\; \|p_r^w - p_h^w\|).
    \label{eq:reward_opp}
\end{equation}
where $\lambda_f>0$ controls the proximity scale. This term encourages a grasp-ready finger configuration without prescribing a hard-coded closure time.

\textbf{Action Penalty.}
Finally, we penalize large end-effector motion to encourage smooth trajectories: $r_{\mathrm{act}} = \|(\Delta p_t, \Delta \omega_t)\|_2^2$.

\subsection{Training Strategy}
\label{sec:curriculum}
To improve training stability and sim-to-real robustness, we combine pose-category curriculum learning with domain randomization. 

\textbf{Pose Curriculum.}
Training uniformly on upright and inverted mugs from the start is unstable due to the substantial geometric difference between the two categories. We therefore use a two-stage sampling schedule over pose categories. Early in training, episodes are sampled from upright and inverted mugs with probabilities $(0.7,0.3)$, allowing the policy to first acquire reliable handle approach and closure behavior. We then switch to a balanced distribution $(0.5,0.5)$ to train a single policy equally across both categories.

\textbf{Domain Randomization.}
\label{sec:dr}
At each episode reset, we randomize mug placement, mug geometry, and pose-estimation noise. Mug positions are sampled within a $20{\times}20$\,cm workspace region, and handle yaw is sampled uniformly from $[-45^\circ,45^\circ]$. Mug geometry is randomized by sampling from a diverse set of mug instances. To model pose-estimation error, we add Gaussian noise to the mug position and orientation, calibrated from the category-level pose estimator CPPF++~\citep{you2024cppf++} evaluated on 100 workspace scenes with ground-truth poses. Full ranges are listed in Appendix~\ref{apx:impl}.

%% file: sections/50exp.tex
\section{Experiments}
\label{sec:exp}

\begin{figure*}[t]
\centering
\includegraphics[width=\textwidth]{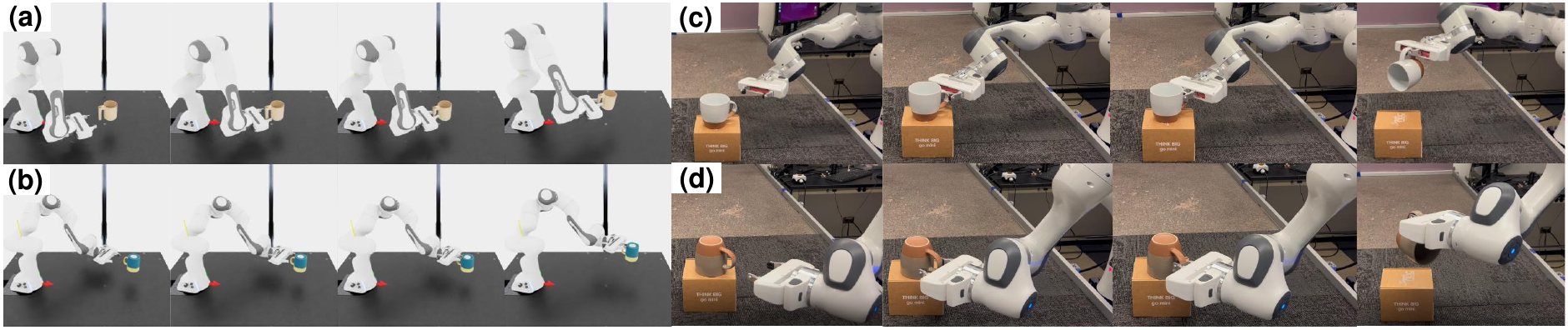}
\caption{
\textbf{Qualitative rollouts of AnyMug.}
Simulation rollouts (a,b) and real-world rollouts (c,d). In both domains, the policy approaches the handle, aligns the gripper, closes around the handle, and lifts the mug to verify grasp stability.
}
\label{fig:qualitative_rollouts}
\end{figure*}

We evaluate AnyMug in simulation and on a real Franka Panda robot. Our experiments answer three questions:
(1)~Can a single policy perform functional handle grasping across diverse mug instances and object poses?
(2)~How does AnyMug compare with open-loop motion planning and learning-based baselines?
(3)~Which components contribute most to AnyMug?

\subsection{Implementation Details}
\label{sec:exp/expsetup}

\textbf{Simulation.} We train and evaluate the policy in Isaac Sim using a Franka Panda manipulator with a two-finger parallel gripper. Policies run at 12\,Hz for 100 control steps, i.e., 8.33\,s per episode. AnyMug is trained with PPO using 512 parallel environments for 50k training steps. The policy uses a CNN encoder for the canonicalized depth image and an MLP action head for 7D end-effector command prediction. Policies are trained with three random seeds to estimate variance. At evaluation time, we test on unseen mug instances with randomized placements and pose categories. Each policy is evaluated on 500 randomized trials per pose category.

\textbf{Real Robot.} We deploy the learned policy zero-shot on a real 7-DoF Franka Panda arm with a two-finger parallel gripper. A side-view RGB camera observes the workspace, and CPPF++~\citep{you2024cppf++} estimates the 6-DoF mug pose. We infer the pose category $c$ by comparing the mug's estimated canonical vertical axis with the world gravity direction. A fixed top-view depth camera provides the depth observation used for canonicalization. We evaluate on 5 physically distinct mugs unseen during training. To keep the real-world budget tractable across all methods and 2 pose categories, we conduct 2 trials per mug per pose category, resulting in 10 upright and 10 inverted real-world trials per method. Mug placements and handle yaws are sampled from the same distribution as the simulation evaluation and re-used across methods to keep comparisons paired.

\textbf{Metrics.}
Figure~\ref{fig:qualitative_rollouts} shows representative rollouts in simulation and on hardware. A trial is successful if the robot grasps the mug by the handle stably and lifts it at least $h_{\min}=10$\,cm above the table. The lift phase serves as a grasp-stability check rather than a separate manipulation objective. 
We report four metrics.
\textbf{Success Rate (SR)}: the fraction of successful trials;
\textbf{Tip/Fall}: the fraction of trials where the mug tips over or falls during approach, grasping, or lifting;
\textbf{Empty}: the fraction of trials where the gripper closes without obtaining a handle grasp;
\textbf{Timeout}: the fraction of trials where the robot fails to close the gripper within the episode horizon.
These metrics capture both task success and failure modes.

\textbf{Baselines.}
We compare AnyMug against one analytical baseline and three learning-based baselines:

\begin{itemize}
    \item \textbf{Motion Planning (MP):} A pose-based baseline that moves the gripper to a predefined grasp pose relative to the estimated mug center. To evaluate robustness to pose errors in simulation, we inject zero-mean Gaussian noise into the grasp offset with $\sigma \in \{0.020,\,0.025,\,0.030\}$\,m. In real-world deployment, MP uses the raw CPPF++ pose estimate without injected noise.

    \item \textbf{Diffusion Policy (DP)~\citep{chi2025diffusion}:} An image-conditioned imitation learning baseline trained on simulation demonstrations from the motion-planning oracle. DP takes the raw top-view depth image and the same proprioceptive state as AnyMug, and predicts end-effector action sequences via iterative denoising. For a fair comparison with AnyMug, DP runs as a single closed-loop policy from the home pose and predicts reaching, alignment, and closure without a motion-planning pre-reaching stage. Hybrid variants that use a planner for coarse approach and DP only for local manipulation could improve absolute success rates, but they are not directly comparable to AnyMug as single end-to-end closed-loop policies.

    \item \textbf{EquiDiff~\citep{wang2024equivariant}:} An image-conditioned diffusion policy baseline that incorporates SO(2) equivariance into the denoising network to improve generalization under planar rotations.

    \item \textbf{EquiBot~\citep{yang2024equibot}:} A point-cloud-based $\mathrm{SIM}(3)$-equivariant diffusion policy designed for data-efficient generalization under changes in scale, rotation, and translation.
\end{itemize}

\subsection{Simulation Results}

\begin{table*}[t]
\centering
\caption{
Simulation results on upright and inverted mugs. We report success rate (SR), mug tipping/falling rate (Tip/Fall), empty grasp rate (Empty), and timeout rate.  All values are percentages. Best results in \textbf{bold}.
}
\label{tab:sim_results}
\resizebox{\linewidth}{!}{
\input{tables/sim_results}
}
\end{table*}

Table~\ref{tab:sim_results} compares AnyMug with the baselines in simulation. AnyMug achieves the highest success rates on both upright and inverted mugs, reaching $93.18\%$ and $94.19\%$ SR, respectively, with near-zero tip/fall and timeout rates. The motion-planning baseline is competitive under small pose perturbations, but degrades sharply as the injected offset noise increases. At $\sigma=0.030$\,m, its SR drops to $62.68\%$ on upright mugs and $59.03\%$ on inverted mugs, with tip/fall rates exceeding $20\%$ in both categories, highlighting the sensitivity of open-loop execution to pose-estimation error. 
Diffusion Policy achieves below $37\%$ SR on both pose categories, with failures dominated by empty grasps and timeouts, indicating difficulty in learning precise handle alignment and closure timing from demonstrations alone. EquiDiff improves over DP, reaching $54.00\%$ SR on upright mugs and $54.6\%$ on inverted mugs, but still remains far below AnyMug. EquiBot achieves only $12.80\%$ SR on upright mugs and $19.35\%$ on inverted mugs, suggesting that equivariance to scale, rotation, and translation alone is insufficient for fine-grained handle-geometry reasoning and functional contact timing.
Overall, these results show that mug-handle grasping requires precise handle localization, gripper alignment, and closure timing, which AnyMug supports by canonicalizing both observations and actions in a mug-centric frame and training with handle-aware closed-loop rewards.

\paragraph{Sim-to-sim transfer.}
To assess cross-backend portability, we deploy the PhysX-trained AnyMug policy in a Newton-compatible Franka setup without finetuning. The policy uses the same canonicalized observation and 7D action interface. As detailed in Appendix~\ref{app:physx_newton_transfer}, the policy maintains high success rates across upright and inverted mugs under pose-estimation noise ($85.2$--$90.0$\% SR), suggesting that the canonicalized observation-action interface transfers across physics backends.

\subsection{Real-World Results}

\begin{table*}[t]
\centering
\caption{
Real-world grasping results on 5 unseen mugs (2 trials per mug per pose category, $n{=}10$ trials per method per pose category). All values are percentages. Best results in \textbf{bold}. Standard deviations are omitted because per-cell sample size ($n{=}10$) is dominated by binomial sampling noise; we instead emphasize relative trends across methods.
}
\label{tab:real_world_results}
\resizebox{\linewidth}{!}{
\input{tables/real_world}
}
\end{table*}

Table~\ref{tab:real_world_results} reports zero-shot real-world performance on 5 unseen physical mugs (2 trials per mug, $n{=}10$ per pose category). Given the small sample size, we focus on relative trends and dominant failure modes. AnyMug transfers directly from simulation without real-world finetuning and attains the highest success rate on both pose categories (8/10 each).
Its failures are split between tip/fall and empty grasps (1/10 each), suggesting that the policy generally approaches the handle safely, while failures arise from handle-alignment errors under real perception noise and contact variation.

Real-world MP uses the raw 6-DoF pose estimated by CPPF++ without injected noise. It reaches 7/10 successes on upright mugs and 6/10 on inverted mugs, with failures mainly due to tip/fall and empty grasps. Although CPPF++ pose noise is calibrated, real deployment introduces additional errors from depth artifacts, calibration drift, and physical mug variation, which reduce the reliability of open-loop execution. In contrast, AnyMug uses closed-loop canonicalized feedback to compensate for residual errors and keeps tip/fall to 1/10 in each pose category.

Diffusion Policy, trained on simulation demonstrations from the motion-planning oracle and evaluated zero-shot on hardware, achieves only 2/10 successes on upright mugs and 1/10 on inverted mugs. Failures are dominated by empty grasps and timeouts, indicating difficulty in learning handle alignment and closure timing. This gap likely stems from the need to learn coarse approach, fine alignment, and closure end-to-end while implicitly covering variation in mug placement, handle yaw, and geometry. These challenges favor AnyMug that factors out pose-induced variation through canonicalization. Additional real-world analysis is provided in Appendix~\ref{apx:real_analysis}.

\subsection{Ablation Study}

Due to space constraints, we move the full ablation table and analysis to Appendix~\ref{apx:ablation}. The key takeaways are: (i) the finger opposition reward is essential, as removing it yields $0\%$ SR and $100\%$ empty grasps on both pose categories; (ii) observation--action canonicalization is the next most important component, with SR drops from $93.18\%$ to $31.53\%$ on upright mugs and from $94.19\%$ to $40.99\%$ on inverted mugs without it, with failures dominated by empty grasps; and (iii) pose-estimation noise injection, mug pose randomization, and the pose-category curriculum each provide consistent additional gains, with the first two providing the largest gains.

%% file: tables/sim_results.tex
\begin{tabular}{lcccccccc}
\toprule
\multirow{2}{*}{Method}
& \multicolumn{4}{c}{Upright Mug}
& \multicolumn{4}{c}{Inverted Mug} \\
\cmidrule(lr){2-5} \cmidrule(lr){6-9}
& SR $\uparrow$ & Tip/Fall $\downarrow$ & Empty $\downarrow$ & Timeout $\downarrow$
& SR $\uparrow$ & Tip/Fall $\downarrow$ & Empty $\downarrow$ & Timeout $\downarrow$ \\
\midrule
AnyMug (Ours)
& \textbf{93.18 $\pm$ 0.59} & \textbf{0.53 $\pm$ 0.09} & \textbf{6.29 $\pm$ 0.63} & \textbf{0.00 $\pm$ 0.00}
& \textbf{94.19 $\pm$ 4.52} & \textbf{0.20 $\pm$ 0.28} & \textbf{5.61 $\pm$ 4.25} & \textbf{0.00 $\pm$ 0.00} \\

MP($\sigma=0.020$)
& 89.68 $\pm$ 2.66 & 6.50 $\pm$ 3.38 & 3.82 $\pm$ 0.73 & 0.00 $\pm$ 0.00
& 85.92 $\pm$ 1.03 & 3.81 $\pm$ 0.51 & 10.27 $\pm$ 0.69 & 0.00 $\pm$ 0.00 \\

MP($\sigma=0.025$)
& 81.63 $\pm$ 4.36 & 10.85 $\pm$ 5.08 & 7.52 $\pm$ 1.13 & 0.00 $\pm$ 0.00
& 76.73 $\pm$ 0.87 & 7.87 $\pm$ 1.64 & 15.40 $\pm$ 1.88 & 0.00 $\pm$ 0.00 \\

MP($\sigma=0.030$)
& 62.68 $\pm$ 2.53 & 26.55 $\pm$ 2.27 & 10.78 $\pm$ 1.40 & 0.00 $\pm$ 0.00
& 59.03 $\pm$ 5.41 & 22.26 $\pm$ 7.84 & 18.70 $\pm$ 2.48 & 0.00 $\pm$ 0.00 \\

DP
& 36.48 $\pm$ 1.02 & 2.87 $\pm$ 1.00 & 48.20 $\pm$ 2.90 & 12.45 $\pm$ 3.59
& 34.35 $\pm$ 6.32 & 3.99 $\pm$ 0.75 & 43.99 $\pm$ 4.81 & 17.67 $\pm$ 6.80 \\

EquiDiff
& $54.00 \pm 4.20$ & $4.40 \pm 5.10$ & $23.70 \pm 0.70$ & $17.90 \pm 0.10$
& $54.60 \pm 3.40$ & $2.20 \pm 2.00$ & $23.20 \pm 1.40$ & $20.00 \pm 2.80$ \\

EquiBot
& 12.80 $\pm$ 2.39 & 33.53 $\pm$ 2.81 & 34.82 $\pm$ 5.37 & 19.05 $\pm$ 4.77
& 19.35 $\pm$ 0.70 & 7.94 $\pm$ 1.96 & 22.72 $\pm$ 2.67 & 50.10 $\pm$ 5.19 \\

\bottomrule
\end{tabular}

%% file: tables/real_world.tex
\begin{tabular}{lcccccccc}
\toprule
\multirow{2}{*}{Method}
& \multicolumn{4}{c}{Upright Mug (\%)}
& \multicolumn{4}{c}{Inverted Mug (\%)} \\
\cmidrule(lr){2-5} \cmidrule(lr){6-9}
& SR $\uparrow$ & Tip/Fall $\downarrow$ & Empty $\downarrow$ & Timeout $\downarrow$
& SR $\uparrow$ & Tip/Fall $\downarrow$ & Empty $\downarrow$ & Timeout $\downarrow$ \\
\midrule
AnyMug (Ours)
& \textbf{80} & \textbf{10} & \textbf{10} & \textbf{0}
& \textbf{80} & \textbf{10} & \textbf{10} & \textbf{0} \\

MP (CPPF++)
& 70 & 20 & 10 & 0
& 60 & 10 & 30 & 0 \\

DP
& 20 & 10 & 60 & 10
& 10 & 10 & 60 & 20 \\




\bottomrule
\end{tabular}

%% file: sections/60limit.tex
\section{Conclusion and Limitations}
\label{sec:conclusion}

We propose AnyMug, a reinforcement learning framework for pose-agnostic mug-handle grasping. AnyMug canonicalizes both the top-view depth observation and the end-effector action in a mug-centric frame, reducing pose-induced variation in perception and control. A handle-aware closed-loop policy then learns instance-level adaptation for handle approach, gripper alignment, and closure timing. Experiments in simulation and zero-shot transfer to a real Franka Panda robot demonstrate robust functional grasping across diverse mug instances and pose configurations.

\paragraph{Limitations.}
First, our experiments focus on a single object category, mugs, and two pose categories, upright and inverted. Extending the framework to broader object classes, additional pose modes such as tilted or side-lying mugs, and downstream tasks such as placing, pouring, or hanging remains future work.

Second, AnyMug assumes a clear top-down depth observation of the mug. In cluttered scenes, under severe occlusion, or with reflective mug surfaces, the depth image and the projected handle direction used for canonicalization may become unreliable. A promising direction is to incorporate additional viewpoints, such as a wrist-mounted camera, multi-view depth sensing, or active perception, to improve handle visibility during closed-loop execution.

%% file: sections/70appen.tex
\section{Related Work}
\label{apx:related}

\subsection{Grasp Pose Detection}
A popular paradigm in robotic grasping is to predict a target grasp pose from visual input. Classical methods~\citep{bicchi2000robotic, miller2004graspit, bohg2013data} use analytical grasp metrics or geometric heuristics over known object models. Data-driven methods predict planar image grasps~\citep{redmon2015real, pinto2016supersizing}, learn grasp quality from synthetic depth~\citep{mahler2016dex, mahler2017dex, mahler2019learning}, or estimate full 6-DoF or 7-DoF grasps from point clouds~\citep{ten2017grasp, yan2018learning, mousavian20196, fang2023anygrasp}.
Although effective for general-purpose grasping, these methods typically execute the predicted grasp open loop. Mug handles are thin, curved, and highly instance-dependent, making them sensitive to small pose errors that can cause empty grasps or collisions. AnyMug instead learns a closed-loop policy that continuously adjusts approach, alignment, and closure from visual and proprioceptive feedback.

\subsection{Closed-Loop Vision-Based Grasping}
Closed-loop grasping methods improve robustness by incorporating visual feedback during execution. Prior work~\citep{levine2018learning, dmitry2018qt, viereck2017learning, zeng2018learning} has learned value functions, grasp success predictors, or visual-based policies from large-scale interaction data. More recent approaches~\citep{chi2025diffusion, hsu2025spot} use diffusion policies to denoise action sequences, or vision-language-action models~\citep{zitkovich2023rt} to leverage large-scale pretraining for robot control.
While these methods can handle diverse object configurations given sufficient data, they typically learn pose and orientation generalization implicitly. For functional mug-handle grasping, this requires the policy to learn distinct approach directions and closure timings across handle orientations and pose categories. AnyMug instead explicitly reduces this variation by canonicalizing both observations and actions into a shared mug-centric frame, enabling a single policy to operate in a consistent handle-centered representation.

\subsection{Canonical Representations for Manipulation}
Canonicalization and equivariance are widely used to reduce variation from object transformations. Canonical coordinate representations have been used for category-level perception and pose estimation~\citep{wang2019normalized}.
In manipulation, $\mathrm{SE}(2)$- and $\mathrm{SO}(2)$-equivariant architectures have improved sample efficiency in Transporter Networks and reinforcement learning policies~\citep{zeng2021transporter, huang2022equivariant, zhu2023robot, wang2022so2, wang2022onrobot}.
Equivariant diffusion policies~\citep{wang2024equivariant, yang2024equibot} further impose $\mathrm{SO}(2)$ or $\mathrm{SIM}(3)$ symmetry on denoising networks for improved geometric generalization.
However, these methods typically canonicalize observations or constrain the network while leaving actions in the robot frame, requiring the policy to learn placement-dependent motor commands. AnyMug instead canonicalizes both the top-view depth observation and end-effector action in a shared mug-centric frame, directly coupling visual normalization with control for handle localization, gripper alignment, and closure timing.

\section{Implementation Details}
\label{apx:impl}

\paragraph{Training Hyperparameters.}
Table~\ref{tab:hparams} lists all hyperparameters used for the AnyMug PPO training.
All experiments use a single NVIDIA RTX 4090 GPU.

\begin{table}[h]
\centering
\caption{PPO training hyperparameters for AnyMug.}
\label{tab:hparams}
\input{tables/ppo_param}
\end{table}

\paragraph{Reward Weights.}
Table~\ref{tab:reward_weights} lists the weights of the reward terms in Eq.~\eqref{eq:reward_total} along with the auxiliary shaping terms used during training.

\begin{table}[h]
\centering
\caption{Reward term weights and decay coefficients for AnyMug.}
\label{tab:reward_weights}
\resizebox{\linewidth}{!}{
\input{tables/reward_weights}
}
\end{table}

\paragraph{Policy Architecture.}
\label{apx:arch}
The policy network processes the $120{\times}120{\times}1$ canonical depth image through a four-layer convolutional encoder with channel widths $[16, 32, 64, 128]$, kernels $[6,4,4,3]$ and strides are $[2,2,2,2]$. Each convolution is followed by a ReLU activation and layer normalization. The resulting $128{\times}6{\times}6$ feature map is average-pooled and flattened to obtain a 128D visual feature. This visual feature is concatenated with the $9$D proprioceptive state vector $x_t = (p_{\mathrm{rel}}, q_{\mathrm{rel}}, c) \in \mathbb{R}^{7+K}$ with $K=2$, where $p_{\mathrm{rel}}\in\mathbb{R}^3$ and $q_{\mathrm{rel}}\in\mathbb{R}^4$ denote the
end-effector pose in the mug-centric frame, and $c\in\{0,1\}^K$ is the pose-category indicator. The concatenated feature is passed through a fully connected trunk with hidden
widths $[256,128,64]$ and ELU activations. The actor head outputs the mean of a 7D Gaussian action distribution,
$a_t^c=(\Delta p_t^c,\Delta\omega_t^c,u_t^g)$. The critic uses the same shared trunk and a separate linear value head to predict $V(s_t)$.

\paragraph{Domain Randomization Details.}
\label{apx:dr}
We randomize three quantities at each episode reset. (i)~\emph{Mug placement.}  At each episode reset, the mug is offset from its nominal position by a uniform planar offset in $[-10, +10]$\,cm in both the $x$ and $y$ axes, and its yaw around the world $z$-axis is sampled uniformly from $\mathcal{U}(-0.25\pi,\,0.25\pi)$.  (ii)~\emph{Mug instance.}  At each episode reset, one mug asset is sampled uniformly from the pool of $30$ training mug USD instances.  (iii)~\emph{Pose-estimation noise.}  At every control step, the simulator's ground-truth mug pose is corrupted with additive Gaussian noise calibrated to a real category-level 6-DoF pose estimator~\citep{you2024cppf++}; statistics are fit on $100$ captured workspace scenes by comparing the estimator's output against ground truth.  The resulting position-noise standard deviations are $\bm{\sigma}_p \approx (2.78, 2.57, 2.59)\,\mathrm{mm}$, with means $\bm{\mu}_p \approx (0.17, -0.23, 0.30)\,\mathrm{mm}$.  Rotational noise is applied in axis-angle form with standard deviations $\bm{\sigma}_R \approx (0.169, 0.064, 0.109)\,\mathrm{rad}$ and means $\bm{\mu}_R \approx (-0.008, 0.017, 0.021)\,\mathrm{rad}$, then converted to a perturbation quaternion that left-multiplies the ground-truth orientation.  We do not currently randomize mass, friction, geometry scale, lighting, or depth-image noise; targeting the pose-estimate channel directly is sufficient for zero-shot transfer in our experiments.

\paragraph{Real-Robot Deployment.}
At deployment, a fixed top-down RealSense D435 above the workspace streams depth at $30$\,fps.  The raw depth frame is cropped and resized to $120{\times}120$ to match the simulator-side input resolution, and the same canonical affine warp (Sec.~\ref{sec:canon}) is applied online using the on-robot pose estimate.  The policy runs at $12$\,Hz, matching its training control rate; end-effector commands are converted to joint targets by a differential inverse-kinematics controller, and the Franka Panda internal controller closes the joint-level loop at $1000$\,Hz.

\section{PhysX-to-Newton Sim-to-Sim Policy Transfer}
\label{app:physx_newton_transfer}

We further evaluate cross-backend portability through a PhysX-to-Newton
sim-to-sim transfer experiment. The same AnyMug checkpoint trained in the
PhysX-based Isaac Lab environment is executed in a Newton-compatible Franka
setup without policy retraining. This experiment tests whether the learned
policy interface, handle-approach behavior, and closure timing remain effective
under a different physics and contact backend.

The policy interface is unchanged during transfer. The policy receives the same
canonicalized depth observation and end-effector state as in PhysX, and outputs
a 7D command: translational displacement, rotational displacement,
and a scalar gripper command. Because the policy operates in end-effector space
rather than joint space, no policy-side joint remapping is
required; backend-specific execution is handled by the Newton-side Franka
controller and contact-evaluation layer.

\paragraph{Newton execution path.}
During Newton execution, the learned policy remains responsible for approaching
the mug handle and deciding when to initiate closure. The end-effector command
predicted in the canonical frame is transformed back to the robot/world frame
and executed by the Newton-compatible Franka controller. When the policy
initiates a grasp attempt near the handle, the rollout enters a contact-gated
execution stage:
\[
\begin{aligned}
&\text{policy-driven approach}
\rightarrow \text{policy-triggered closure}
\rightarrow \text{two-sided contact verification} \\
&\rightarrow \text{staged grasp stabilization}
\rightarrow \text{lift verification}.
\end{aligned}
\]
A grasp is accepted only after two-sided physics contact is established
between the gripper pads and the mug handle. To obtain stable and interpretable
contact behavior across physics backends, the Newton evaluation uses simplified
collision proxies for the finger pads and mug handle. These proxies are used
only in the sim-to-sim transfer test and do not modify the learned policy. After
two-sided contact is verified, the rollout enters a standardized staged lift,
which checks whether the accepted contact configuration supports a successful
handle grasp. Thus, the experiment evaluates cross-backend transfer of the
policy's approach and contact-acquisition behavior, while the post-contact lift
is standardized for consistent Newton evaluation.

\paragraph{Randomization and perturbation.}
Newton evaluation uses randomized mug position and yaw within each pose category,
following the same evaluation protocol as the main simulation experiments. We additionally report results
under three controlled reset-configuration perturbation levels,
$\sigma \in \{0.020, 0.025, 0.030\}$ m, applied to the reset inverse-kinematics
height used to initialize the Newton rollout. These perturbations are applied in
addition to randomized mug placement and yaw, and are not injected into the
learned policy's output action.

\paragraph{Evaluation metrics.}
A trial is successful if the mug is grasped by the handle, lifted at
least 10\,cm above its initial support height, and held for 1\,s after contact
verification. \textbf{Tip/Fall} records trials where the mug tips, falls, or
moves into an invalid pose before success. \textbf{Empty} records trials where
the policy attempts closure but does not obtain a stable contact-verified handle
grasp. \textbf{Timeout} records trials where the task is not completed within
the episode horizon. These metrics separate unsafe object disturbance from
handle-acquisition failures and from horizon-limited executions.

\begin{table}[t]
\centering
\caption{PhysX-to-Newton contact-gated sim-to-sim transfer under increasing grasp-target perturbation. The same PhysX-trained AnyMug checkpoint is evaluated in the Newton-compatible execution path without retraining. A grasp is evaluated only after two-sided handle contact is verified. All values are percentages.}
\label{tab:physx_newton_contact_transfer}
\begin{tabular}{lcccc}
\toprule
Noise level & SR $\uparrow$ & Tip/Fall $\downarrow$ & Empty $\downarrow$ & Timeout $\downarrow$ \\
\midrule
\multicolumn{5}{c}{\textbf{Upright Mug}} \\
\midrule
$\sigma = 0.020$
& $89.2 \pm 1.4$ & $0.0 \pm 0.0$ & $10.8 \pm 1.4$ & $0.0 \pm 0.0$ \\

$\sigma = 0.025$
& $85.4 \pm 1.6$ & $0.0 \pm 0.0$ & $14.2 \pm 1.6$ & $0.4 \pm 0.3$ \\

$\sigma = 0.030$
& $90.0 \pm 1.3$ & $0.0 \pm 0.0$ & $9.6 \pm 1.3$ & $0.4 \pm 0.3$ \\

\midrule
\multicolumn{5}{c}{\textbf{Inverted Mug}} \\
\midrule
$\sigma = 0.020$
& $87.2 \pm 1.5$ & $0.0 \pm 0.0$ & $12.8 \pm 1.5$ & $0.0 \pm 0.0$ \\

$\sigma = 0.025$
& $87.8 \pm 1.5$ & $0.0 \pm 0.0$ & $12.2 \pm 1.5$ & $0.0 \pm 0.0$ \\

$\sigma = 0.030$
& $85.2 \pm 1.6$ & $0.0 \pm 0.0$ & $13.8 \pm 1.5$ & $1.0 \pm 0.4$ \\
\bottomrule
\end{tabular}
\end{table}

\paragraph{Results and interpretation.}
Table~\ref{tab:physx_newton_contact_transfer} shows that the PhysX-trained
AnyMug policy preserves strong handle-grasping performance when transferred to
the Newton-compatible execution path. Across all perturbation levels, the
Tip/Fall rate remains near zero, suggesting that the transferred policy does not
generally destabilize the mug during approach. The dominant failure mode is
empty grasping, indicating that the remaining sim-to-sim gap is primarily due to
contact acquisition and handle alignment rather than unsafe approach motion.

The results also show that the policy remains robust under increasing
grasp-target perturbation. Even at the largest perturbation level
\(\sigma = 0.030\) m, the policy maintains high success rates for both upright
and inverted mugs. This supports the claim that observation-action
canonicalization transfers across physics backends at the policy-interface
level: the policy continues to approach the handle from useful directions and
initiates closure near feasible grasp configurations, while the Newton-side
contact verification determines whether the grasp is physically accepted.

\section{Ablation Study}
\label{apx:ablation}

\begin{table*}[t]
\centering
\caption{
Ablation study in simulation. All values are percentages. Best results in \textbf{bold}.
}
\label{tab:sim_ablation}
\resizebox{\linewidth}{!}{
\input{tables/sim_ablation}
}
\end{table*}

Table~\ref{tab:sim_ablation} isolates the contribution of each component in AnyMug. Removing the finger opposition reward leads to complete failure on both pose categories, with $0\%$ SR and $100\%$ empty grasps. This confirms that reaching and orientation rewards alone do not provide sufficient signal for functional handle grasping, while the opposition reward is essential for encouraging the fingers to wrap around the handle before closure.

Canonicalization is also critical. Without observation-action canonicalization, SR decreases from $93.18\%$ to $31.53\%$ on upright mugs and from $94.19\%$ to $40.99\%$ on inverted mugs. The resulting failures are overwhelmingly empty grasps, showing that the shared mug-centric representation is key for consistent handle localization and closure across different placements and handle orientations.

The remaining ablations highlight the importance of robust training. Removing pose-estimation noise reduces SR to $35.71\%$ on upright mugs and yields high variance on inverted mugs ($69.11\%\pm35.37$), indicating sensitivity to imperfect canonical frames. Removing mug pose randomization also substantially degrades performance, reducing SR to $56.45\%$ and $58.75\%$ on upright and inverted mugs. Finally, removing the curriculum causes a smaller but consistent drop to $81.14\%$ and $85.81\%$. Overall, the ablations show that finger opposition reward and canonicalization are the primary drivers of functional handle grasping, while pose estimation noise, mug pose randomization, and curriculum learning improve robustness and training stability.

\section{Additional Real-World Analysis}
\label{apx:real_analysis}

Table~\ref{tab:real_world_results} reports real-world results with $n{=}10$ trials per method and pose category. Because this sample size yields wide binomial confidence intervals, we use the real-world experiments primarily to compare consistent trends and dominant failure modes across methods, rather than to emphasize small percentage-point differences. For example, 8/10 successes corresponds to a 95\% Wilson interval of approximately $[49\%,94\%]$. The load-bearing evidence is therefore the consistent ordering across methods, pose categories, and failure metrics.

\paragraph{AnyMug.}
AnyMug achieves 8/10 successes on both upright and inverted mugs without real-world finetuning. The two failures in each pose category are split between tip/fall and empty grasp. This indicates that the policy usually reaches the handle without destabilizing the object, while the remaining failures are primarily due to residual handle-alignment errors under real depth artifacts, pose-estimation noise, and contact variation. Compared with simulation, the real-world success rate drops by roughly 10--15 percentage points, which is expected given visual-domain shift, calibration error, and physical mug variation.

\paragraph{Motion planning.}
In the real-world experiments, MP uses the raw 6-DoF pose estimated by CPPF++ without additional injected noise. It achieves 7/10 successes on upright mugs and 6/10 on inverted mugs, with failures mainly due to tip/fall and empty grasps. This is consistent with the simulation sweep in Table~\ref{tab:sim_results}, where open-loop MP degrades as pose-offset noise increases. Although the calibrated CPPF++ position noise (in Appendix~\ref{apx:dr}) is smaller than the simulated perturbations, real-world execution introduces additional errors from depth artifacts, gripper-to-base calibration drift, contact mismatch, and physical mug-geometry variation. These factors compound pose error and make open-loop execution brittle. In contrast, AnyMug uses closed-loop canonicalized feedback to refine its approach and closure online, achieving higher success while keeping tip/fall to 1/10 in each pose category.

\paragraph{Diffusion Policy.}
Diffusion Policy, trained on motion-planning-oracle demonstrations in simulation and evaluated zero-shot on hardware, achieves only 2/10 (upright) and 1/10 (inverted), with failures dominated by empty grasps (6/10 on each pose) and timeouts (1--2/10): the policy reaches the workspace but rarely aligns the gripper with the handle or commits to closure in time. Two factors compound this gap. First, DP runs end-to-end from the home pose with no motion-planning pre-reaching, so its representational capacity must cover coarse approach, fine handle alignment, and closure timing. Second, the demonstration distribution must implicitly span the full variation in mug placement, handle yaw, and instance geometry sampled at evaluation, which is difficult to cover densely from a tractable demonstration budget. Both factors favor methods that decouple pose-induced variation from instance-level execution through canonicalization, as AnyMug does.

\paragraph{Interpretation.}
Overall, the real-world results support the importance of explicitly canonicalizing both perception and control. Motion planning relies on accurate pose estimates and cannot correct residual errors during execution, while end-to-end imitation must learn coarse approach, fine handle alignment, and closure timing jointly from limited demonstrations. AnyMug reduces this burden by expressing both observations and actions in a mug-centric canonical frame, allowing the closed-loop policy to focus on instance-specific handle geometry and contact timing.





%% file: tables/ppo_param.tex
\begin{tabular}{lc}
\toprule
Hyperparameter & Value \\
\midrule
Number of parallel environments ($N$)    & 512 \\
Total training timesteps                 & 50{,}000 \\
Episode length                           & $8.33$\,s \\
Control rate                             & $12$\,Hz (physics $120$\,Hz, decimation $10$) \\
PPO Rollout length per environment           & 64 steps \\
PPO epochs per iteration ($M$)           & 8 \\
Mini-batches per epoch                   & 8 \\
Discount factor ($\gamma$)               & 0.99 \\
GAE parameter ($\lambda_{\mathrm{GAE}}$) & 0.95 \\
PPO clip coefficient ($\epsilon$)        & 0.2 \\
Value clip coefficient                   & 0.2 \\
Learning rate ($\eta$)                   & $1\times10^{-4}$ \\
Learning-rate scheduler                  & KL-adaptive, threshold $0.01$ \\
Entropy coefficient                      & 0.0 \\
Value loss coefficient                   & 2.0 \\
Max gradient norm                        & 1.0 \\
Optimizer                                & Adam \\
\bottomrule
\end{tabular}

%% file: tables/reward_weights.tex
\begin{tabular}{llc}
\toprule
Term & Description & Value \\
\midrule
$w_p$, $\lambda_p$ & Reaching reward weight, decay   & $5.0$, $10.0$ \\
$w_R$, $\lambda_R$ & Orientation reward weight, decay & $5.0$, $2.0$ \\
$w_o$              & Finger-opposition reward weight  & $5.0$  \\
$w_a$              & Action penalty weight (applied to 6-DoF EE delta only; gripper-command dim excluded) & $50.0$ \\
\midrule
$w_f$, $\lambda_f$ & Per-finger gated handle-distance reward (aux.) & $5.0$, $10.0$ \\
$w_b$              & Finger-below-table penalty (aux.) & $5.0$ \\
$w_{ec}$           & Early-close penalty (aux., active when gripper closes far from handle) & enabled \\
\bottomrule
\end{tabular}

%% file: tables/sim_ablation.tex
\begin{tabular}{lcccccccc}
\toprule
\multirow{2}{*}{Method}
& \multicolumn{4}{c}{Upright Mug}
& \multicolumn{4}{c}{Inverted Mug} \\
\cmidrule(lr){2-5} \cmidrule(lr){6-9}
& SR $\uparrow$ & Tip/Fall $\downarrow$ & Empty $\downarrow$ & Timeout $\downarrow$
& SR $\uparrow$ & Tip/Fall $\downarrow$ & Empty $\downarrow$ & Timeout $\downarrow$ \\
\midrule
AnyMug (Ours)
& \textbf{93.18 $\pm$ 0.59} & \textbf{0.53 $\pm$ 0.09} & \textbf{6.29 $\pm$ 0.63} & \textbf{0.00 $\pm$ 0.00}
& \textbf{94.19 $\pm$ 4.52} & \textbf{0.20 $\pm$ 0.28} & \textbf{5.61 $\pm$ 4.25} & \textbf{0.00 $\pm$ 0.00} \\

w/o Canonicalization
& 31.53 $\pm$ 13.11 & 1.59 $\pm$ 1.19 & 66.88 $\pm$ 14.30 & 0.00 $\pm$ 0.00
& 40.99 $\pm$ 15.59 & 0.79 $\pm$ 0.01 & 58.22 $\pm$ 15.58 & 0.00 $\pm$ 0.00 \\

w/o Finger Opposition Reward
& 0.00 $\pm$ 0.00 & 0.00 $\pm$ 0.00 & 100.00 $\pm$ 0.00 & 0.00 $\pm$ 0.00
& 0.00 $\pm$ 0.00 & 0.00 $\pm$ 0.00 & 100.00 $\pm$ 0.00 & 0.00 $\pm$ 0.00 \\

w/o Pose Estimation Noise
& 35.71 $\pm$ 4.10 & 9.20 $\pm$ 10.38 & 55.09 $\pm$ 14.62 & 0.00 $\pm$ 0.00
& 69.11 $\pm$ 35.37 & 0.49 $\pm$ 0.42 & 30.40 $\pm$ 35.79 & 0.00 $\pm$ 0.00 \\

w/o Mug Pose Randomization
& 56.45 $\pm$ 23.63 & 3.17 $\pm$ 1.41 & 40.38 $\pm$ 25.04 & 0.00 $\pm$ 0.00
& 58.75 $\pm$ 14.66 & 0.80 $\pm$ 0.60 & 40.16 $\pm$ 15.55 & 0.00 $\pm$ 0.00 \\

w/o Curriculum
& 81.14 $\pm$ 3.43 & 1.86 $\pm$ 0.96 & 17.00 $\pm$ 2.48 & 0.00 $\pm$ 0.00
& 85.81 $\pm$ 4.75 & 0.39 $\pm$ 0.55 & 13.80 $\pm$ 4.20 & 0.00 $\pm$ 0.00 \\

\bottomrule
\end{tabular}

%% file: root_arxiv.bbl
\begin{thebibliography}{38}
\providecommand{\natexlab}[1]{#1}
\providecommand{\url}[1]{\texttt{#1}}
\expandafter\ifx\csname urlstyle\endcsname\relax
  \providecommand{\doi}[1]{doi: #1}\else
  \providecommand{\doi}{doi: \begingroup \urlstyle{rm}\Url}\fi

\bibitem[Bicchi and Kumar(2000)]{bicchi2000robotic}
A.~Bicchi and V.~Kumar.
\newblock Robotic grasping and contact: A review.
\newblock In \emph{Proceedings 2000 ICRA. Millennium conference. IEEE international conference on robotics and automation. Symposia proceedings (Cat. No. 00CH37065)}, volume~1, pages 348--353. IEEE, 2000.

\bibitem[Bohg et~al.(2013)Bohg, Morales, Asfour, and Kragic]{bohg2013data}
J.~Bohg, A.~Morales, T.~Asfour, and D.~Kragic.
\newblock Data-driven grasp synthesis—a survey.
\newblock \emph{IEEE Transactions on robotics}, 30\penalty0 (2):\penalty0 289--309, 2013.

\bibitem[Sun et~al.(2025)Sun, Curtis, You, Xu, Koehle, Chen, Huang, Guibas, Chitta, Schwager, et~al.]{sun2024arch}
J.~Sun, A.~Curtis, Y.~You, Y.~Xu, M.~Koehle, Q.~Chen, S.~Huang, L.~Guibas, S.~Chitta, M.~Schwager, et~al.
\newblock Arch: Hierarchical hybrid learning for long-horizon contact-rich robotic assembly.
\newblock \emph{CoRL}, 2025.

\bibitem[Chen et~al.(2020)Chen, Ye, Sun, Fan, Hu, Wang, and Lu]{chen2020transferable}
X.~Chen, Z.~Ye, J.~Sun, Y.~Fan, F.~Hu, C.~Wang, and C.~Lu.
\newblock Transferable active grasping and real embodied dataset.
\newblock In \emph{2020 IEEE International Conference on Robotics and Automation (ICRA)}, pages 3611--3618. IEEE, 2020.

\bibitem[Chen et~al.(2026)Chen, Zheng, Yu, Huang, Sun, Goldberg, Wen, Abbeel, Shentu, Wu, et~al.]{chen2026sarm2}
Q.~Chen, H.~Zheng, J.~Yu, S.~Huang, J.~Sun, K.~Goldberg, C.~Wen, P.~Abbeel, Y.~Shentu, P.~Wu, et~al.
\newblock Sarm2: Multi-task stage aware reward modeling for self improving robotic manipulation.
\newblock \emph{arXiv preprint arXiv:2606.10305}, 2026.

\bibitem[Huang et~al.(2026)Huang, Shao, Wang, Chen, Sun, Guo, Schwager, and Bohg]{huang2026breaking}
S.~Huang, J.~Shao, K.~Wang, Q.~Chen, J.~Sun, Y.~Guo, M.~Schwager, and J.~Bohg.
\newblock Breaking lock-in: Preserving steerability under low-data vla post-training.
\newblock \emph{arXiv preprint arXiv:2604.23121}, 2026.

\bibitem[Huang et~al.(2025)Huang, Davies, Yan, Sun, Chen, and Hu]{huang2025spatial}
Y.~Huang, T.~Davies, J.~Yan, J.~Sun, X.~Chen, and L.~Hu.
\newblock Spatial robograsp: Generalized robotic grasping control policy.
\newblock \emph{arXiv preprint arXiv:2505.20814}, 2025.

\bibitem[Wang et~al.(2023)Wang, Fan, Sun, Zhang, Fei-Fei, Xu, Zhu, and Anandkumar]{wang2023mimicplay}
C.~Wang, L.~Fan, J.~Sun, R.~Zhang, L.~Fei-Fei, D.~Xu, Y.~Zhu, and A.~Anandkumar.
\newblock Mimicplay: Long-horizon imitation learning by watching human play.
\newblock \emph{CoRL}, 2023.

\bibitem[Huang et~al.(2025)Huang, Chen, Zhang, Sun, and Schwager]{huang2025particleformer}
S.~Huang, Q.~Chen, X.~Zhang, J.~Sun, and M.~Schwager.
\newblock Particleformer: A 3d point cloud world model for multi-object, multi-material robotic manipulation.
\newblock \emph{CoRL}, 2025.

\bibitem[Mahler et~al.(2017)Mahler, Liang, Niyaz, Laskey, Doan, Liu, Ojea, and Goldberg]{mahler2017dex}
J.~Mahler, J.~Liang, S.~Niyaz, M.~Laskey, R.~Doan, X.~Liu, J.~A. Ojea, and K.~Goldberg.
\newblock Dex-net 2.0: Deep learning to plan robust grasps with synthetic point clouds and analytic grasp metrics.
\newblock \emph{arXiv preprint arXiv:1703.09312}, 2017.

\bibitem[Mousavian et~al.(2019)Mousavian, Eppner, and Fox]{mousavian20196}
A.~Mousavian, C.~Eppner, and D.~Fox.
\newblock 6-dof graspnet: Variational grasp generation for object manipulation.
\newblock In \emph{Proceedings of the IEEE/CVF international conference on computer vision}, pages 2901--2910, 2019.

\bibitem[Fang et~al.(2023)Fang, Wang, Fang, Gou, Liu, Yan, Liu, Xie, and Lu]{fang2023anygrasp}
H.-S. Fang, C.~Wang, H.~Fang, M.~Gou, J.~Liu, H.~Yan, W.~Liu, Y.~Xie, and C.~Lu.
\newblock Anygrasp: Robust and efficient grasp perception in spatial and temporal domains.
\newblock \emph{IEEE Transactions on Robotics}, 39\penalty0 (5):\penalty0 3929--3945, 2023.

\bibitem[Qi et~al.(2017)Qi, Su, Mo, and Guibas]{qi2017pointnet}
C.~R. Qi, H.~Su, K.~Mo, and L.~J. Guibas.
\newblock Pointnet: Deep learning on point sets for 3d classification and segmentation.
\newblock In \emph{Proceedings of the IEEE conference on computer vision and pattern recognition}, pages 652--660, 2017.

\bibitem[You et~al.(2024)You, He, Liu, Xiong, Wang, and Lu]{you2024cppf++}
Y.~You, W.~He, J.~Liu, H.~Xiong, W.~Wang, and C.~Lu.
\newblock Cppf++: Uncertainty-aware sim2real object pose estimation by vote aggregation.
\newblock \emph{IEEE Transactions on Pattern Analysis and Machine Intelligence}, 46\penalty0 (12):\penalty0 9239--9254, 2024.

\bibitem[Levine et~al.(2018)Levine, Pastor, Krizhevsky, Ibarz, and Quillen]{levine2018learning}
S.~Levine, P.~Pastor, A.~Krizhevsky, J.~Ibarz, and D.~Quillen.
\newblock Learning hand-eye coordination for robotic grasping with deep learning and large-scale data collection.
\newblock \emph{The International journal of robotics research}, 37\penalty0 (4-5):\penalty0 421--436, 2018.

\bibitem[Dmitry et~al.(2018)Dmitry, Alex, Peter, Julian, Alexander, Eric, Deirdre, Ethan, Mrinal, Vincent, et~al.]{dmitry2018qt}
K.~Dmitry, I.~Alex, P.~Peter, I.~Julian, H.~Alexander, J.~Eric, Q.~Deirdre, H.~Ethan, K.~Mrinal, V.~Vincent, et~al.
\newblock Qt-opt. scalable deep reinforcement learning for vision-based robotic manipulation.
\newblock \emph{arXiv preprint}, 2018.

\bibitem[Chi et~al.(2025)Chi, Xu, Feng, Cousineau, Du, Burchfiel, Tedrake, and Song]{chi2025diffusion}
C.~Chi, Z.~Xu, S.~Feng, E.~Cousineau, Y.~Du, B.~Burchfiel, R.~Tedrake, and S.~Song.
\newblock Diffusion policy: Visuomotor policy learning via action diffusion.
\newblock \emph{The International Journal of Robotics Research}, 44\penalty0 (10-11):\penalty0 1684--1704, 2025.

\bibitem[Zitkovich et~al.(2023)Zitkovich, Yu, Xu, Xu, Xiao, Xia, Wu, Wohlhart, Welker, Wahid, et~al.]{zitkovich2023rt}
B.~Zitkovich, T.~Yu, S.~Xu, P.~Xu, T.~Xiao, F.~Xia, J.~Wu, P.~Wohlhart, S.~Welker, A.~Wahid, et~al.
\newblock Rt-2: Vision-language-action models transfer web knowledge to robotic control.
\newblock In \emph{Conference on Robot Learning}, pages 2165--2183. PMLR, 2023.

\bibitem[Makoviychuk et~al.(2021)Makoviychuk, Wawrzyniak, Guo, Lu, Storey, Macklin, Hoeller, Rudin, Allshire, Handa, et~al.]{makoviychuk2021isaac}
V.~Makoviychuk, L.~Wawrzyniak, Y.~Guo, M.~Lu, K.~Storey, M.~Macklin, D.~Hoeller, N.~Rudin, A.~Allshire, A.~Handa, et~al.
\newblock Isaac gym: High performance gpu-based physics simulation for robot learning.
\newblock \emph{arXiv preprint arXiv:2108.10470}, 2021.

\bibitem[Schulman et~al.(2017)Schulman, Wolski, Dhariwal, Radford, and Klimov]{schulman2017proximal}
J.~Schulman, F.~Wolski, P.~Dhariwal, A.~Radford, and O.~Klimov.
\newblock Proximal policy optimization algorithms.
\newblock \emph{arXiv preprint arXiv:1707.06347}, 2017.

\bibitem[Wang et~al.(2024)Wang, Hart, Surovik, Kelestemur, Huang, Zhao, Yeatman, Wang, Walters, and Platt]{wang2024equivariant}
D.~Wang, S.~Hart, D.~Surovik, T.~Kelestemur, H.~Huang, H.~Zhao, M.~Yeatman, J.~Wang, R.~Walters, and R.~Platt.
\newblock Equivariant diffusion policy.
\newblock In \emph{Conference on Robot Learning (CoRL)}, 2024.

\bibitem[Yang et~al.(2024)Yang, Cao, Deng, Antonova, Song, and Bohg]{yang2024equibot}
J.~Yang, Z.-a. Cao, C.~Deng, R.~Antonova, S.~Song, and J.~Bohg.
\newblock {EquiBot}: {SIM}(3)-equivariant diffusion policy for generalizable and data-efficient learning.
\newblock In \emph{Conference on Robot Learning (CoRL)}, 2024.

\bibitem[Miller and Allen(2004)]{miller2004graspit}
A.~T. Miller and P.~K. Allen.
\newblock Graspit! a versatile simulator for robotic grasping.
\newblock \emph{IEEE Robotics \& Automation Magazine}, 11\penalty0 (4):\penalty0 110--122, 2004.

\bibitem[Redmon and Angelova(2015)]{redmon2015real}
J.~Redmon and A.~Angelova.
\newblock Real-time grasp detection using convolutional neural networks.
\newblock In \emph{2015 IEEE international conference on robotics and automation (ICRA)}, pages 1316--1322. IEEE, 2015.

\bibitem[Pinto and Gupta(2016)]{pinto2016supersizing}
L.~Pinto and A.~Gupta.
\newblock Supersizing self-supervision: Learning to grasp from 50k tries and 700 robot hours.
\newblock In \emph{2016 IEEE international conference on robotics and automation (ICRA)}, pages 3406--3413. IEEE, 2016.

\bibitem[Mahler et~al.(2016)Mahler, Pokorny, Hou, Roderick, Laskey, Aubry, Kohlhoff, Kr{\"o}ger, Kuffner, and Goldberg]{mahler2016dex}
J.~Mahler, F.~T. Pokorny, B.~Hou, M.~Roderick, M.~Laskey, M.~Aubry, K.~Kohlhoff, T.~Kr{\"o}ger, J.~Kuffner, and K.~Goldberg.
\newblock Dex-net 1.0: A cloud-based network of 3d objects for robust grasp planning using a multi-armed bandit model with correlated rewards.
\newblock In \emph{2016 IEEE international conference on robotics and automation (ICRA)}, pages 1957--1964. IEEE, 2016.

\bibitem[Mahler et~al.(2019)Mahler, Matl, Satish, Danielczuk, DeRose, McKinley, and Goldberg]{mahler2019learning}
J.~Mahler, M.~Matl, V.~Satish, M.~Danielczuk, B.~DeRose, S.~McKinley, and K.~Goldberg.
\newblock Learning ambidextrous robot grasping policies.
\newblock \emph{Science robotics}, 4\penalty0 (26):\penalty0 eaau4984, 2019.

\bibitem[Ten~Pas et~al.(2017)Ten~Pas, Gualtieri, Saenko, and Platt]{ten2017grasp}
A.~Ten~Pas, M.~Gualtieri, K.~Saenko, and R.~Platt.
\newblock Grasp pose detection in point clouds.
\newblock \emph{The International Journal of Robotics Research}, 36\penalty0 (13-14):\penalty0 1455--1473, 2017.

\bibitem[Yan et~al.(2018)Yan, Hsu, Khansari, Bai, Pathak, Gupta, Davidson, and Lee]{yan2018learning}
X.~Yan, J.~Hsu, M.~Khansari, Y.~Bai, A.~Pathak, A.~Gupta, J.~Davidson, and H.~Lee.
\newblock Learning 6-dof grasping interaction via deep geometry-aware 3d representations.
\newblock In \emph{2018 IEEE International Conference on Robotics and Automation (ICRA)}, pages 3766--3773. IEEE, 2018.

\bibitem[Viereck et~al.(2017)Viereck, Pas, Saenko, and Platt]{viereck2017learning}
U.~Viereck, A.~Pas, K.~Saenko, and R.~Platt.
\newblock Learning a visuomotor controller for real world robotic grasping using simulated depth images.
\newblock In \emph{Conference on robot learning}, pages 291--300. PMLR, 2017.

\bibitem[Zeng et~al.(2018)Zeng, Song, Welker, Lee, Rodriguez, and Funkhouser]{zeng2018learning}
A.~Zeng, S.~Song, S.~Welker, J.~Lee, A.~Rodriguez, and T.~Funkhouser.
\newblock Learning synergies between pushing and grasping with self-supervised deep reinforcement learning.
\newblock In \emph{2018 IEEE/RSJ International Conference on Intelligent Robots and Systems (IROS)}, pages 4238--4245. IEEE, 2018.

\bibitem[Hsu et~al.(2025)Hsu, Wen, Xu, Narang, Wang, Zhu, Biswas, and Birchfield]{hsu2025spot}
C.-C. Hsu, B.~Wen, J.~Xu, Y.~Narang, X.~Wang, Y.~Zhu, J.~Biswas, and S.~Birchfield.
\newblock Spot: Se (3) pose trajectory diffusion for object-centric manipulation.
\newblock In \emph{2025 IEEE International Conference on Robotics and Automation (ICRA)}, pages 4853--4860. IEEE, 2025.

\bibitem[Wang et~al.(2019)Wang, Sridhar, Huang, Valentin, Song, and Guibas]{wang2019normalized}
H.~Wang, S.~Sridhar, J.~Huang, J.~Valentin, S.~Song, and L.~J. Guibas.
\newblock Normalized object coordinate space for category-level 6d object pose and size estimation.
\newblock In \emph{Proceedings of the IEEE/CVF conference on computer vision and pattern recognition}, pages 2642--2651, 2019.

\bibitem[Zeng et~al.(2021)Zeng, Florence, Tompson, Welker, Chien, Attarian, Armstrong, Krasin, Duong, Sindhwani, et~al.]{zeng2021transporter}
A.~Zeng, P.~Florence, J.~Tompson, S.~Welker, J.~Chien, M.~Attarian, T.~Armstrong, I.~Krasin, D.~Duong, V.~Sindhwani, et~al.
\newblock Transporter networks: Rearranging the visual world for robotic manipulation.
\newblock In \emph{Conference on Robot Learning}, pages 726--747. PMLR, 2021.

\bibitem[Huang et~al.(2022)Huang, Wang, Walters, and Platt]{huang2022equivariant}
H.~Huang, D.~Wang, R.~Walters, and R.~Platt.
\newblock Equivariant transporter network.
\newblock \emph{arXiv preprint arXiv:2202.09400}, 2022.

\bibitem[Zhu et~al.(2023)Zhu, Wang, Su, Biza, Walters, and Platt]{zhu2023robot}
X.~Zhu, D.~Wang, G.~Su, O.~Biza, R.~Walters, and R.~Platt.
\newblock On robot grasp learning using equivariant models.
\newblock \emph{Autonomous Robots}, 47\penalty0 (8):\penalty0 1175--1193, 2023.

\bibitem[Wang et~al.(2022{\natexlab{a}})Wang, Walters, and Platt]{wang2022so2}
D.~Wang, R.~Walters, and R.~Platt.
\newblock {SO}(2)-equivariant reinforcement learning.
\newblock In \emph{International Conference on Learning Representations (ICLR)}, 2022{\natexlab{a}}.

\bibitem[Wang et~al.(2022{\natexlab{b}})Wang, Jia, Zhu, Walters, and Platt]{wang2022onrobot}
D.~Wang, M.~Jia, X.~Zhu, R.~Walters, and R.~Platt.
\newblock On-robot learning with equivariant models.
\newblock In \emph{Conference on Robot Learning (CoRL)}, 2022{\natexlab{b}}.

\end{thebibliography}
